\documentclass[journal]{IEEEtran}
\ifCLASSINFOpdf
\else
\fi
\usepackage{cite}
\usepackage{amsmath,amssymb,amsfonts}
\usepackage{algorithmic}
\usepackage{graphicx}
\usepackage{subcaption}
\usepackage{textcomp}
\usepackage{wrapfig}
\usepackage{algorithm}
\usepackage{lipsum}
\usepackage[hidelinks]{hyperref}
\usepackage{xcolor}
\usepackage{tabularx,booktabs}
\newcolumntype{C}{>{\centering\arraybackslash}X} 
\def\BibTeX{{\rm B\kern-.05em{\sc i\kern-.025em b}\kern-.08em
    T\kern-.1667em\lower.7ex\hbox{E}\kern-.125emX}}
\begin{document}
%
\title{Fail-Safe Human Detection for Drones Using a Multi-Modal Curriculum Learning Approach} 
%
%
%

\author{Ali Safa, \IEEEmembership{Student Member, IEEE}, 
Tim Verbelen, Ilja Ocket, \IEEEmembership{Member, IEEE}, 
\\
Andr\'e Bourdoux, \IEEEmembership{Senior Member, IEEE},
Francky Catthoor, \IEEEmembership{Fellow, IEEE}, 
Georges G.E. Gielen, \IEEEmembership{Fellow, IEEE}

\thanks{Ali Safa, Ilja Ocket, Francky Catthoor and Georges G.E Gielen are with imec and the Department of Electrical Engineering, KU Leuven, 3001 Leuven, Belgium (e-mail: Ali.Safa@imec.be; Ilja.Ocket@imec.be; Francky.Catthoor@imec.be; Georges.Gielen@kuleuven.be).}
\thanks{Tim Verbelen is with imec and the Department of Electrical Engineering, U Gent, 9000 Gent, Belgium (e-mail: Tim.Verbelen@imec.be).}
\thanks{Andr\'e Bourdoux is with imec, 3001 Leuven, Belgium (e-mail: Andre.Bourdoux@imec.be).}
}
%
%

\markboth{}%
{Safa \MakeLowercase{\textit{et al.}}: }
%



\maketitle

\begin{abstract}
Drones are currently being explored for safety-critical applications where human agents are expected to evolve in their vicinity. In such applications, robust people avoidance must be provided by fusing a number of sensing modalities in order to avoid collisions. Currently however, people detection systems used on drones are solely based on standard cameras besides an emerging number of works discussing the fusion of imaging and event-based cameras. On the other hand, radar-based systems provide up-most robustness towards environmental conditions but do not provide complete information on their own and have mainly been investigated in automotive contexts, not for drones. In order to enable the fusion of radars with both event-based and standard cameras, we present KUL-UAVSAFE, a first-of-its-kind dataset for the study of safety-critical people detection by drones. In addition, we propose a baseline CNN architecture with cross-fusion highways and introduce a curriculum learning strategy for multi-modal data termed SAUL, which greatly enhances the robustness of the system towards hard RGB failures and provides a significant gain of 15\% in peak $\boldsymbol{F_1}$ score compared to the use of \textit{BlackIn}, previously proposed for cross-fusion networks. We demonstrate the real-time performance and feasibility of the approach by implementing the system in an edge-computing unit. We release our dataset and additional material in the project home page.    
\end{abstract}

\begin{IEEEkeywords}
People detection, sensor fusion, UAVs, curriculum learning
\end{IEEEkeywords}

\section*{Supplementary Material}
\noindent
Please visit the project home page for the dataset and additional material at: \url{https://ali20480.github.io/PhDPortfolio/kuluavsafe/}

%
\IEEEpeerreviewmaketitle

\section{Introduction}
\label{sec:introduction}
\IEEEPARstart{H}{uman} detection for small \textit{Unmanned Aerial Vehicles} (UAVs) or drones is of up-most importance in safety-critical applications where human agents and drones are expected to evolve side by side. Such use cases range from automated logistics and warehouse inspection to people search and rescue. The performance of a human detection system for drones can be characterized by the four following performance factors. The first performance factor (PF1) is the \textit{detection accuracy and recall}, measured using \textit{precision-recall} curves \cite{averageprec}, by comparing the bounding boxes returned by the detection system against the ground truth bounding boxes. The second performance factor (PF2) is the \textit{detection speed}, measured in \textit{frames per second} (FPS), which determines how fast the drone can react to a human entering or evolving in its field of view. The third performance factor (PF3) is the \textit{energy and area consumption} of the detection system (drones have limited battery budgets): can the system be deployed on small, low-power edge processors such as a Google Coral edge-TPU \cite{edgeTPU}? The fourth performance factor (PF4) is the \textit{detection robustness} to hard sensor faults such as complete RGB blackout, and to environmental conditions such as e.g., dirt and water droplets on camera lenses, low light and so on. 

Thanks to the enormous progress in \textit{deep learning}, high-precision and high-recall object detection (PF1) using standard imaging cameras is nowadays ubiquitous. This has mainly been enabled by Deep Neural Networks (DNNs) based on the \textit{You Only Look Once} (YOLO) principle \cite{YOLOv4}, that achieve good 
performance 
with a high inference speed (PF2. $\sim 65$ FPS) mainly on bulky, power-hungry desktop GPUs \cite{YOLOv4} or on lower-power embedded GPU platforms such as the NVIDIA Jetson Nano ($\sim 30$ FPS for a Tiny-YOLO \cite{tinyYOLO}) however, still unsuited for the tightest of energy budgets 
($\sim 10$W vs. $\sim 2$W for an edge-TPU \cite{edgeTPU}). On the other hand, deploying those DNNs on small, ultra-low-power edge processors (PF3) for fast embedded inference is actively being investigated by studying pruning, quantization and network design strategies, tightly coupled with DNN hardware accelerator co-design (see \cite{embeddedDNNbook} for a comprehensive overview). Among those works, the \textit{SqueezeNet} architecture has been proposed in \cite{squeezenet} as an AlexNet-level accuracy backbone network with $<50$ MB model size, making it well-suited for fast, ultra-low-power edge inference ($\sim 2$ms inference time in an edge-TPU \cite{edgeTPU}). For this reason, we adopt \textit{SqueezeNet} \cite{squeezenet} as our backbone network throughout this paper. 

Regarding PF4, system robustness to sensor failure and environmental conditions has been investigated by fusing standard imaging cameras with other sensing modalities such as radar \cite{homography1} or Dynamic Vision Sensors \cite{eventbasedsurvey} (DVS) (event-based cameras), which asynchronously report the change in per-pixel brightness as a train of binary pulses with $\sim 1 \mu$s resolution. They provide a significantly higher dynamic range than standard cameras ($\sim 140$dB vs. $\sim 60$dB) \cite{eventbasedsurvey}. Thus, fusing a DVS sensor with a standard imaging sensor as done in a DAVIS camera \cite{DAVIS} enables high-speed and accurate sensing, even in low-light conditions \cite{eventbasedsurvey}. On the other hand, using a DAVIS camera alone still has drawbacks regarding PF4. Sensing in darkness or extreme low light is still difficult, dirt in the lens can still create occlusions, and so on. In addition, the sensor does not provide ranging information on its own and must be processed through rather compute-expensive depth estimation algorithms in order to provide ranging \cite{depthest}. For those reasons, we choose to fuse a DAVIS camera with a \textit{Frequency Modulated Continuous Wave} (FMCW) radar in this work. Indeed, a radar intrinsically provides ranging information, is very robust towards environmental effects, can sense in total blackout and is insensitive to occlusion by dirt \cite{throughwall}. In addition, it must be noted that compute-expensive depth estimation must preferably be performed when a sensor fails. 

Within this context, the novel contributions of this work are the following. We acquire a first-of-its-kind dataset termed KUL-UAVSAFE, providing RGB, DVS, radar, accelerometer, gyroscope and altitude modalities in an indoor industrial environment with a small drone flying near a walking person. Second, we propose a novel baseline \textit{Convolutional Neural Network} (CNN) for human detection based on a SqueezeNet backbone \cite{squeezenet}, augmented with \textit{cross-fusion} highways \cite{origCrossfusion} for the DVS and radar modalities, resulting in an end-to-end, monolithic network that can learn \textit{at which resolution stage} fusion should happen \cite{origCrossfusion} 
Finally, we report our system performance and robustness in different sensor ablation contexts by introducing a new multi-modal \textit{curriculum learning} \cite{curr} procedure for our cross-fusion network that we call \textit{\underline{S}h\underline{a}ke-\underline{u}p \underline{L}earning} (SAUL). Compared to the \textit{BlackIn} procedure previously introduced in \cite{origCrossfusion}, it provides an average gain of $15\%$ on the peak $F_1$ score of the detection system. We have deployed the system on embedded hardware and demonstrate the real-time feasibility of the approach.


This paper is organized as follows. Related works are discussed in section \ref{s1}. Our methods are introduced in section \ref{s2}. Results and their discussion are presented in section \ref{s3}. Conclusions are provided in section \ref{s4}.

\section{Related Works}
\label{s1}
In recent years, a number of datasets featuring DVS cameras for UAV and for object detection have been proposed \cite{dronedvs1, dronedvs2, dronedvs3} but none of them contain radar data. Yet, radar has extensively been used in automotive applications to increase the system robustness and thus, its safety \cite{origCrossfusion, modularfusion}. A number of automotive datasets fusing a standard imaging camera with a FMCW radar have been proposed in literature \cite{nuScenes, carrada}, none of them featuring DVS data. In contrast, our KUL-UAVSAFE dataset is the first one fusing RGB, DVS and radar modalities for assessing the safety of UAVs towards human agents in indoor environments.

RGB-radar fusion through \textit{deep neural network} (DNN) processing has already been studied in a number of works, using different fusion strategies. Among those, the authors in \cite{homography1} proposed a road target classification and tracking system using a 79-GHz FMCW radar and a standard imaging camera. They adopt a late fusion approach, applying object recognition independently to the camera data using a YOLOv3 detector \cite{yolov3}, and to the radar data, using a CNN-LSTM network. Then the detections from both modalities are projected on a common plane by homography, and an extended Kalman filter is used for data fusion and tracking. While their system achieves a very high performance even when an RGB failure is simulated, their system is most suited for a static setup, requiring a power-hungry desktop GPU in order to meet real-time speed. Compared to our work, they use two distinct DNNs (one for each modality), each requiring GPU compute power, while we use a single CNN based on the SqueezeNet backbone \cite{squeezenet}. This drastically reduces the memory footprint and the required computing power, which enables edge inference for drones on a low-power CNN accelerator \cite{edgeTPU}. The authors in \cite{origCrossfusion} proposed a cross-fusion network to classify objects in an automotive context with radar and camera data from the nuScenes dataset \cite{nuScenes}. They clearly demonstrate that a cross-fusion approach is beneficial as it enables their network to learn \textit{where} fusion should happen the most. Inspired by the architecture proposed in \cite{origCrossfusion}, we also make use of a cross-fusion approach in our CNN design (although we use a totally different architecture, better suited for edge computing). In addition, a learning approach called \textit{BlackIn} is proposed in \cite{origCrossfusion}, where the RGB data is randomly blacked out during training with a probability rate of $0.2$. This prevents the cross-fusion network proposed in \cite{origCrossfusion} to focus learning on the RGB modality only and provides a better detection performance. Still, \cite{origCrossfusion} does not provide an RGB sensor ablation study and it is therefore unclear whether their system is robust towards a \textit{hard} camera failure. In contrast, we demonstrate in this paper that combining the \textit{BlackIn} procedure of \cite{origCrossfusion} with a modified \textit{curriculum learning} procedure \cite{curr} enables our cross-fusion network to significantly outperform the performance obtained via \textit{BlackIn} only, even when hard sensor failures are simulated. 
\section{Sensor suite and data acquisition}
\label{s2}
In this section, we first briefly introduce the sensing principle of the DVS camera and the FMCW radar used in our RGB-DVS-radar fusion setup. Then, we describe the integration of each sensor with the drone. Finally, we present the dataset acquisition.
\subsection{Background Sensor Theory}
\subsubsection{DVS principle}
Compared to standard imaging shutters, an event-based camera outputs binary events of polarity $p_k$ for each independent pixel $\vec{x}_k$ in an asynchronous manner whenever the change in log-intensity at time $t_k$, $\Delta L(\vec{x}_k,t_k)$, crosses a certain threshold $C$ \cite{eventbasedsurvey}. The polarity $p_k$ is positive if $\Delta L(\vec{x}_k,t_k)$ is increasing and \textit{vice versa} for $\Delta L(\vec{x}_k,t_k)$ decreasing. Using the usual \textit{optical flow constraint}, it can be shown that events are generated by moving edges \cite{eventbasedsurvey} (since edges have high contrast $|\Delta L(\vec{x}_k,t_k)|$). Therefore, DVS data does not convey the same information as RGB data, which justifies the use of a \textit{cross-fusion} approach as this latter enables the CNN to learn at which \textit{feature scale} RGB-DVS fusion should happen the most \cite{origCrossfusion}.
\subsubsection{FMCW Radar principle} A FMCW radar emits chirps
\begin{equation}
    p_q(t) = \exp{j(2 \pi f_c t + \pi \alpha t^2)}
    \label{chirp}
\end{equation}
at its transmit antenna (where $q$ is the chirp index, $f_c$ is the start frequency and $\alpha$ is the slope), and senses the reflected waves through a receive antenna array, providing not only ranging but also angle-of-arrival (AoA) information \cite{mydetector}. The received signal at each receive antenna is demodulated and the following signal is obtained \cite{myrad}:
\begin{equation}
    r_{qm}(t) = \sum_{i=1}^{N_{t}} \xi_i e^{j(-2 \pi \alpha T_{d_i}  t - 2 \pi f_c T_{d_i}  + \pi \alpha T_{d_i}^2 + m\phi_i)}
    \label{eq2}
\end{equation}
where $m$ is the receive antenna index, $N_t$ is the number of reflecting targets, $\phi_i$ is the phase shift of target $i$ due to its AoA, and $T_{d_i}$ is the round-trip time from the target $i$ to the radar antenna, linked to the distance $d_i$ between the radar and the target following $T_{d_i} = \frac{2d_i}{c}$ (with $c$ the speed of light). Through successive FFT processing steps, a range-Doppler-azimuth (RDA) heatmap can be obtained for each radar frame (a packet of $N_s$ chirps) \cite{myrad}. Peaks in the RDA heatmap indicate targets at certain ranges and AoAs, and with a certain radial (or Doppler) velocities. A \textit{Constant False Alarm Rate} (CFAR) detector (which adapts its threshold depending on the local noise estimate) can then be used to detect potential targets against noise in the RDA heatmaps \cite{mydetector}. As for DVS data, radar data does not convey the same amount of information as RGB data. Indeed, compared to an isolated RGB pixel, a single radar pixel can convey much richer information such as range, radial velocity or AoA, which, again, justifies the use of a \textit{cross-fusion} approach \cite{origCrossfusion}.

\begin{table}[htbp]
\begin{center}
\begin{tabular}{|c|c|c|c|c|c|}
\hline
$\theta_{res}$ &$d_{res}$ & $d_{max}$  &  $v_{min}$ & $v_{res}$ & $v_{max}$\\
\hline
$15^\circ$ & $17.8$ cm & $9.11$ m & $0.36$ m/s & $0.12$ m/s & $3.84$ m/s\\
\hline
\end{tabular}
\caption{\textit{\textbf{Radar parameters.} $\theta_{res}$ is the AoA resolution, $d_{res}$ is the range resolution, $d_{max}$ is the maximum range that can be sensed, $v_{min}$ is the Doppler velocity threshold, $v_{res}$ is the velocity resolution, and $v_{max}$ is the maximum velocity that can be sensed.}}
\label{radarparam}
\end{center}
\end{table}
\subsection{Sensor suite description}
\begin{figure}[htbp]
\centerline{\includegraphics[scale=0.5]{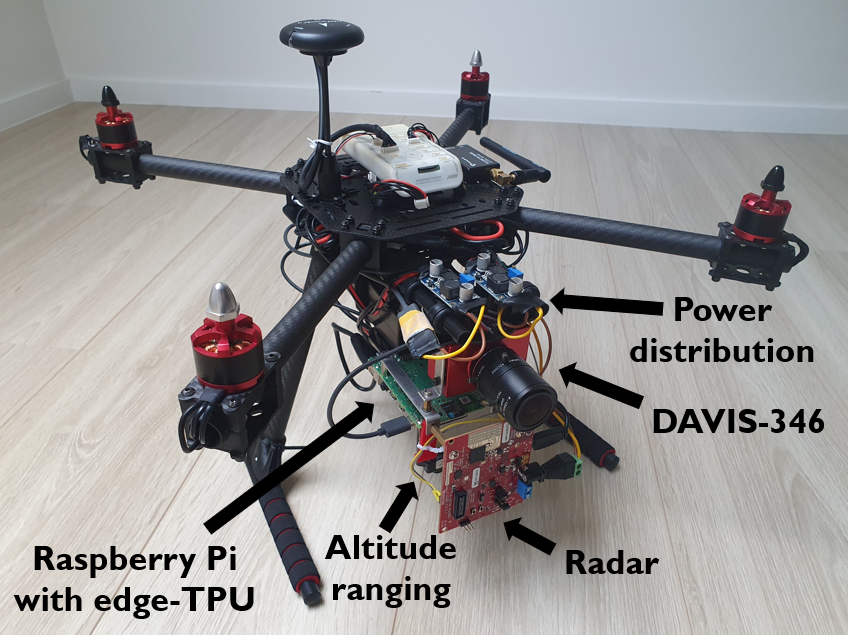}}
\caption{\textit{\textbf{Our drone setup} with the sensor suite mounted on it, as indicated with the arrows.}}
\label{figdr}
\end{figure}

Our first-of-its-kind sensor suite is composed of a DAVIS-346 camera \cite{DAVIS} simultaneously providing RGB frames at 30 FPS and event data with $1\mu$s resolution. The DAVIS-346 also provides accelerometer and gyroscope data that we log as well during the flights. In addition, we use an AWR1443 79-GHz radar, which provides radar detection frames at 30 FPS. In contrast to a number of works that first acquire the raw radar ADC data and post-process them in an external computing unit \cite{radaradc2, radaradc3}, we perform radar detection directly in the embedded radar chip MCU using the detector proposed in \cite{mydetector}. By performing detection in the embedded radar MCU rather than in an external computing unit, we 1) drastically reduce the data bandwidth between the radar and the central computing unit, 2) remove the need for additional FPGA-based modules (e.g., DCA1000EVM) used to translate radar signals to gigabit Ethernet, making our setup less bulky, significantly cheaper (saving $\sim500$\$) and much less energy consuming (saving $>3$W) compared to setups using those FPGA-based modules. Finally, we use a TF-MINI range sensor to log the drone altitude at $\sim 20$ FPS. The radar is time-synchronized with the DAVIS-346 camera by sending a pulse signal to the \textit{synch input} of the camera when the radar acquisition cycle starts. Finally, a Raspberry Pi 4 is used to log the data coming from all sensors. We mount this sensor suite on a \textit{NXP HoverGames} drone platform. Fig. \ref{figdr} shows the complete drone-sensor setup.
\begin{table}[htbp]
\begin{tabularx}{0.49\textwidth}{@{}l*{1}{C}c@{}}
\toprule
Name  & Background difficulty & Number of captures \\ 
\midrule
\textbf{1)} h-wall   & easy      &     2 \\ 
\textbf{2)} h-benches  & medium        &     2 \\ 
\textbf{3)} h-shelves  & medium        &       2 \\ 
\textbf{4)} h-aisles  & challenging         &     7  \\ 

\midrule
\textbf{5)} f-wall   & easy      &      4\\ 
\textbf{6)} f-shelves-benches  & medium        &        2\\ 
\textbf{7)} f-gate      & medium       &     2\\ 
\textbf{8)} f-aisles      & challenging       &        4\\ 
\textbf{9)} f-aisles-gate      & challenging       &     2\\ 
\textbf{10)} f-aisles-wall      & challenging       &     3\\ 

\bottomrule
\end{tabularx}
\caption{\textit{\textbf{Hover (h) and fly (f) sets} with \textbf{6} different walking subjects (30 acquisitions in total).}}
\label{acqs}
\end{table}
\begin{figure}[!t]
\centerline{\includegraphics[scale=0.55]{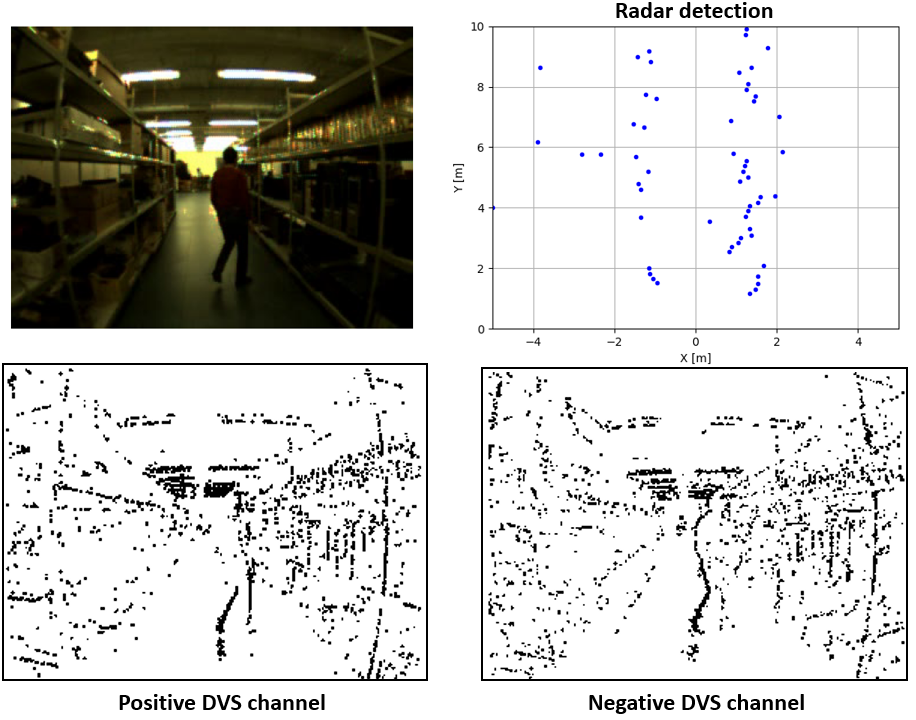}}
\caption{\textit{\textbf{Snippet from the \textit{f-aisles-1} acquisition.} The static background has a non-zero Doppler velocity with regard to the drone as the drone is flying forward. Detections from the right and left shelves are thus returned and can be used for obstacle avoidance.}}
\label{movfig}
\end{figure}
\begin{figure*}[!t]
\centering
\captionsetup{justification=centering}
    \includegraphics[scale = 0.65]{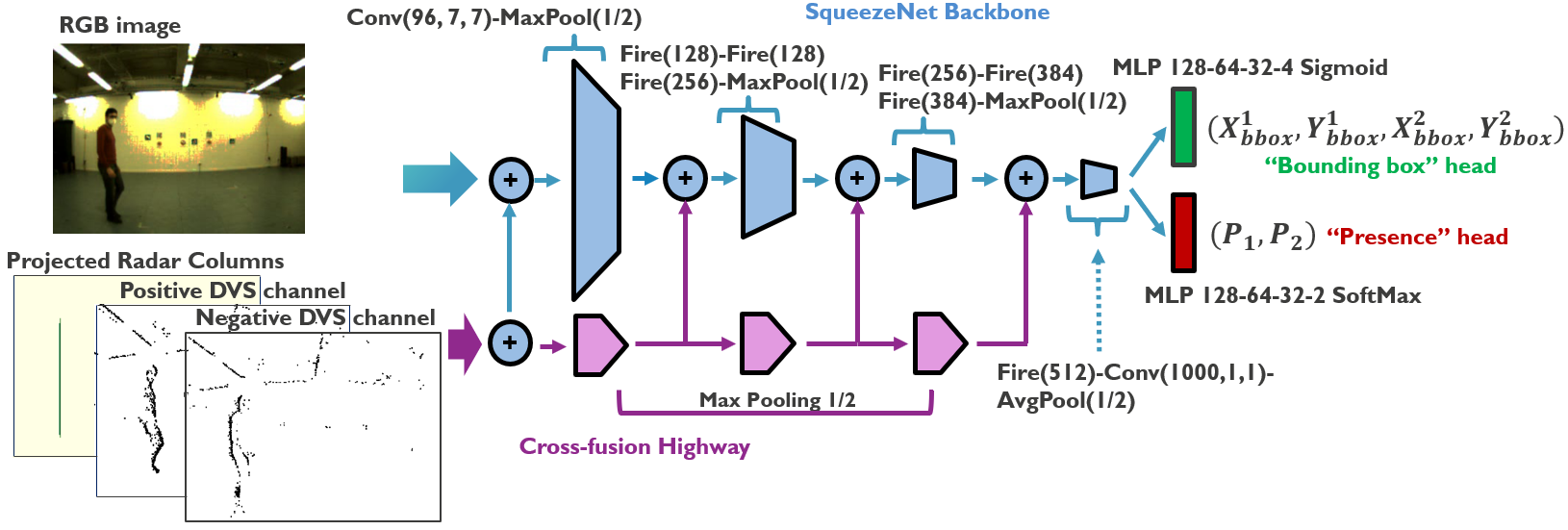}
    \caption{\textit{\textbf{Our CNN architecture with cross-fusion highways}. The RGB, DVS and radar inputs are concatenated and fed to the input of the SqueezeNet backbone. In addition, the DVS and radar inputs are concatenated and fed to the input of the cross-fusion highway. During training, the network learns at \textit{which scale} fusion must happen the most along the backbone. Finally, two MLP heads are used for bounding box regression and for human presence detection.}}
    \label{baseCNN}
\end{figure*}
\subsection{Dataset acquisition}
With this drone-sensor setup, we have acquired a dataset termed KUL-UAVSAFE, intended for the study of robust people detection algorithms, in order to provide safety in environments where humans and drones evolve side by side. As the data acquisition was conducted \textit{indoor} during the COVID-19 pandemic, the dataset currently contains a single walking person only (our dataset will be extended with multiple walking people in the future). The data is captured in different locations with different backgrounds, from simple (e.g., \textit{wall} background) to more challenging (e.g., \textit{aisles} background). The dataset features 6 distinct human subjects in order to capture variations in clothes color, shape, walking style and so on. In all acquisitions, the human subject is asked to walk arbitrarily in order to capture a large number of situations (human crossing, coming close, turning, and so on). The drone is either set to hover with light movements (to make the data more challenging) or to fly straight. The radar parameters are given in Table \ref{radarparam}. Tables \ref{acqs} gives an overview of the acquired data. 
During all our acquisitions, we reject detections having a Doppler velocity magnitude smaller than a threshold $v_{min}$. This enables an automatic, on-line rejection of the static background obstacles when the drone is in a (quasi) hover state, which only requires the sensing of moving agents that could collide with the hovering drone (as static obstacles cannot collide with a static drone). On the other hand, this filtering still allows the detection of static obstacles when the drone is moving (see Fig. \ref{movfig}), as the background has a non-zero Doppler velocity with regard to the drone in this case (moving obstacles with a velocity close to the drone velocity are also detected thanks to the fine-grain velocity resolution of the radar). Finally, we refer the reader to the project home page for additional videos showcasing the dataset.
\section{Baseline CNN design}
This section presents the CNN architecture we use (see Fig. \ref{baseCNN}). We first detail the input data pre-processing. Then, we describe our baseline CNN architecture. 
Finally, we introduce our novel SAUL training strategy. 
\label{s2bis}

\subsection{Input pre-processing}
The input to our CNN is a concatenation of the RGB, DVS and radar modalities, forming a $(356\times260\times6)$ tensor. The first three channels are the RGB channels, rescaled between $0$ and $1$. The fourth and fifth channels correspond to the positive and the negative polarity of DVS frames obtained by taking the mean of the event time stamps $t_{k,n}$ for each pixel $\vec{x}_k$ during a time window $\Delta T = 10$ ms, as follows:
\begin{equation}
    D(\vec{x}_k, t_0) = \frac{\sum_{n=1}^{N_k(\Delta T )} t_{k,n} - t_0}{N_k(\Delta T )}
    \label{meanacc}
\end{equation}
Here, $D$ denotes the final pixel value of the DVS frame, $N_k(\Delta T )$ is the number of events generated at pixel location $\vec{x}_k$ during $\Delta T$ and $t_0$ is the initial time of the accumulation window. Eq. (\ref{meanacc}) is applied separately to the positive and the negative channels, obtaining one DVS frame for each polarity. Then, the DVS frames are normalized between $0$ and $1$ and concatenated with the RGB. Finally, the radar data must be projected from the range-azimuth view to the same perspective domain as the RGB and DVS images. As the drone altitude is not fixed, computing a homography between the radar range-azimuth plane and the cameras directly as in \cite{homography1} cannot be done since the radar is not a \textit{projective} sensor. This means that, as the drone altitude changes (for fixed $X,Y$ location), the radar detection map does not change as the radar cannot distinguish the objects elevation extents. 

\begin{figure}[htbp]
\centerline{\includegraphics[scale=0.2]{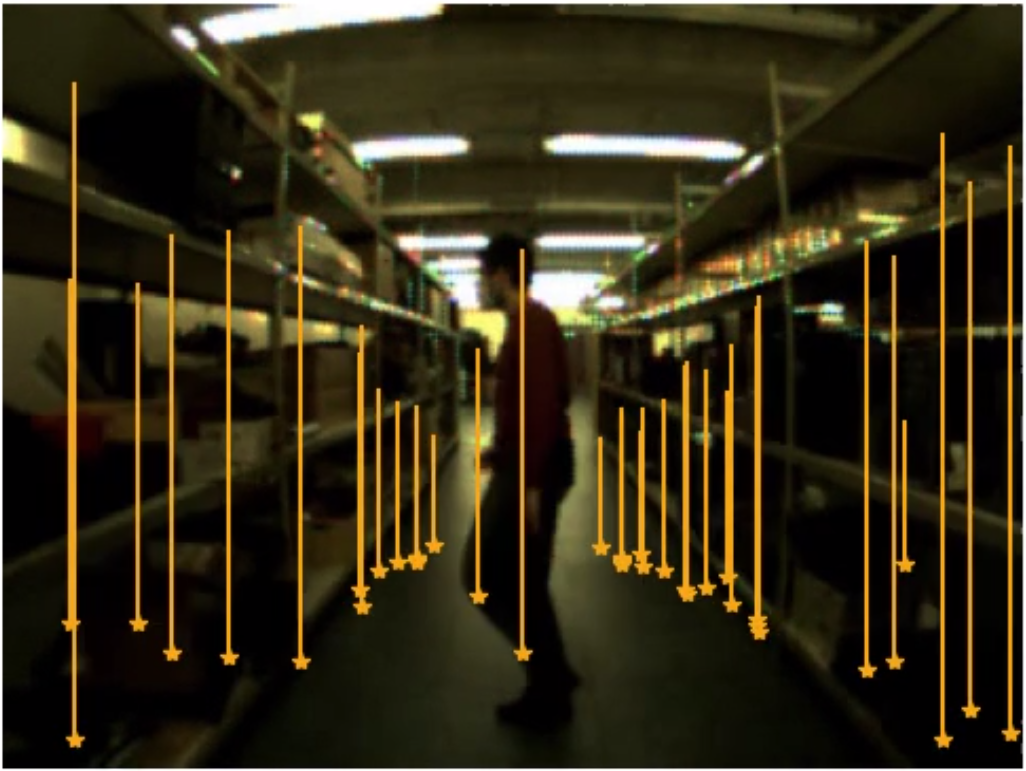}}
\caption{\textit{\textbf{Radar columns} obtained by projecting the range-azimuth radar detections on the image plane by homography (star-shaped markers) and extending them in height to cover their spatial context.}}
\label{figcolum}
\end{figure}

Instead, we use the TF-MINI range sensor to measure the drone altitude $h$ and we augment the radar range-azimuth detections $(r_i, \theta_i)$ with this elevation information to obtain tuples of the form $(r_i, \theta_i, -h)$ for each detection $i$. Thus, we assume that the detections are generated from objects that are anchored to the ground, which is a very generic assumption (shelves, benches and walking humans are all tied to the ground floor). Using these augmented radar detections, we construct an intermediate image-like representation of the radar data as follows:

\begin{equation}
    \begin{bmatrix}
    x_i\\
    y_i
    \end{bmatrix} = \frac{f}{-h} 
    \begin{bmatrix}
    r_i \cos{\theta_i} \\
    r_i \sin{\theta_i}
    \end{bmatrix}
    \label{intermediaterep}
\end{equation}
where $(x_i, y_i)$ represent the coordinate of the radar detection $i$ in the intermediate space and $f$ is a focal length factor, which we chose equal to the focal length of our DAVIS-356 camera (measured through checkerboard calibration). Then it is sufficient to compute the homography matrix $M$ using more than eight detection coordinates $(x_i, y_i)$ and their corresponding locations on the image planes, for different drone heights. We perform this step offline as an extrinsic calibration procedure using a corner reflector as in \cite{homography1}. Then, we project the intermediate point coordinates $(x_i, y_i)$ on the image plane, and we extend their height vertically as done in \cite{origCrossfusion} to better cover their spatial context, assuming a prior height of $1.5$ m for all detections. We set the pixel values of the resulting \textit{radar columns} as the normalized depth value $d_i = \frac{r_i \sin{\theta_i}}{d_{max}}$ measured by the radar. Fig. \ref{figcolum} shows an example of the \textit{projected radar columns} obtained through this process (input to the CNN). It must be noted that slight calibration errors may occur. But, as our cross-fusion CNN can learn to mitigate them during training, we did not observe any problem linked to such errors during our experiments.   
Finally, we concatenate the \textit{radar columns} with the RGB and DVS channels as the last channel of the input tensors $T_n$:
\begin{equation}
    T_n = (I_n^{3},D_n^{2},R_n^{1})
    \label{tenssse}
\end{equation}
where $n$ is the frame index, $I_n^{3}$ is the RGB image, $D_n^{2}$ is the DVS frame with its two polarity channels, and $R_n^{1}$ is the single radar channel. In order to obtain training labels, we use a YOLOv3 detector \cite{yolov3} to automatically estimate the bounding box coordinates and the presence or absence of a human subject. In the case of occlusion, the YOLOv3 still returns a bounding box covering the visible part of the walking subject. The bounding box coordinates are encoded as:  
\begin{equation}
   \underline{b} = (X^1_{bbox}, Y^1_{bbox}, X^2_{bbox}, Y^2_{bbox})
   \label{bboxlab}
\end{equation}
where $X^1_{bbox}, Y^1_{bbox}$ is the normalized top-left coordinate and $X^2_{bbox}, Y^2_{bbox}$ is the normalized bottom-right coordinate. The normalization is done by dividing the coordinates by the length of their respective axis to obtain values between $0$ and $1$. Finally, we encode the presence or the absence of a human subject as a one-hot encoded vector of two elements $(P_1,P_2)$. Naturally, it should be noted that the YOLOv3 annotation is not perfect but still, the learning is very robust towards annotation noise since the number of mistakes in annotation is very small compared to the correct annotations. The successive tensors and their corresponding labels are saved at the frame rate of the RGB camera ($30$ FPS) to form the final labelled datasets for each acquisition. 

\subsection{CNN design}
\label{cnnarch}
Fig. \ref{baseCNN} shows the baseline CNN architecture used in this work. The network backbone is a SqueezeNet \cite{squeezenet}, chosen for its advantageous balance between high accuracy, low inference speed on typical edge-TPUs ($\sim 2$ ms) \cite{edgeTPU} and low memory footprint ($<50$MB). This performance is achieved thanks to the use of \textit{Fire} modules instead of the typical \textit{Convolutional} layers (see Fig. \ref{baseCNN}) \cite{squeezenet}. We augment the SqueezeNet backbone by adding a \textit{cross-fusion highway} \cite{origCrossfusion}, which successively applies \textit{Max Pooling} on its input data and concatenates the intermediate results with the corresponding feature maps of the backbone, for each resolution level (see Fig. \ref{baseCNN}). In contrast to traditional \textit{early fusion} which only fuses data at maximal resolution, our \textit{cross-fusion} setting enables the network to learn \textit{at which resolution level} fusion must best take place \cite{origCrossfusion} (see ablation experiments in section \ref{crossvsearly}). Finally, we connect two distinct \textit{Multi-Layer Perceptron} (MLP) heads to the backbone. The \textit{bounding box} head (see the green box in Fig. \ref{baseCNN}) features four sigmoid outputs that are used for the regression of the normalized bounding box coordinates (\ref{bboxlab}). The \textit{presence} head (see the red box in Fig. \ref{baseCNN}) features a two-class \textit{SoftMax} output for predicting the presence or the absence of the human subject. The input to the SqueezeNet backbone is the concatenation of the RGB, DVS and radar maps (\ref{tenssse}), while the input to the cross-fusion highway is the concatenation of the DVS and the radar maps $(D_n^{2}, R_n^{1})$ only. As our dataset is currently composed of a single walking subject and as our aim is to analyse the sensor fusion encoding and learning strategies, we used a \textit{single} bounding box regression output at the head of our CNN. Thus, our backbone can naturally be extended to a \textit{multi} bounding box case by adding e.g., the output layers of a Tiny-YOLO \cite{tinyYOLO} or of a \textit{RetinaNet} \cite{focaloss}, as done in \cite{origCrossfusion}.

\subsection{Choice of the loss function}
In a general way, we can write the CNN learning objective as the minimization of the following loss function:

\begin{equation}
     \mathcal{L}_{tot} = \lambda \mathcal{L}_{bbox} + (1-\lambda) \mathcal{L}_{pres}
     \label{eqloss}
 \end{equation}
where $\mathcal{L}_{bbox}$ is the bounding box loss, measuring the distance between the predicted bounding box and the ground truth, $\mathcal{L}_{pres}$ is the presence loss associated with the human presence detection head in fig. \ref{baseCNN} and $\lambda$ is a hyper-parameter controlling the balance between the two losses. As usual for classification, we choose $\mathcal{L}_{pres}$ to be the \textit{cross-entropy} loss. For bounding box regression, the learning task is to minimize the distance $\mathcal{L}_{bbox}(\hat{\underline{b}}, \underline{b})$ between the estimated top-left and bottom-right coordinates of the bounding box $\hat{\underline{b}}$ in Fig. \ref{baseCNN} and the ground-truth coordinates $\underline{b}$ (\ref{bboxlab}). In a number of works \cite{ssd}, $\mathcal{L}_{bbox}$ is chosen as the \textit{Huber} (or \textit{smooth} $L_1$) loss for its robustness towards outlier coordinates. Instead, we use the inverted Huber (or \textit{berHu}) loss (\ref{berhu}) in order to focus learning on the difficult examples \cite{berhu}. During our experiments, this choice was confirmed by observing that the \textit{berHu} loss gave significantly better results than the \textit{Huber} loss, qualitatively leading to \textit{less fuzzy behaviour} in the bounding box estimates during testing. 
\begin{equation}
    \mathcal{L}_{berHu}(\hat{b}_i,b_i) = \begin{cases} |\hat{b}_i - b_i|, & \mbox{if } |\hat{b}_i - b_i| \leq c \\ \frac{(\hat{b}_i - b_i)^2 + c^2}{2c}, & \mbox{else}\end{cases}
    \label{berhu}
\end{equation}
Similar to \cite{berhu}, we adaptively set the $c$ parameter of (\ref{berhu}) as $c = 0.2 \times \max_{i} |\hat{b}_i - b_i|$ where the $i$ index denotes the $i^{th}$ element of the bounding box coordinate vector. Regarding the setting of the hyper-parameter $\lambda$ in (\ref{eqloss}), we observed during our experiments that we systematically achieved better results by first training our CNN for bounding box regression only and then, freezing the weights of the network, adding a second MLP head for human presence detection (see Fig. \ref{baseCNN}) and training this second head independently. Indeed, optimizing for the detection and bounding box regression objectives at the same time may introduce antagonistic effects for our specific architecture, which can harm learning. Instead, training for the bounding boxes first enables the backbone CNN to learn robust human feature representations, well-suited for the presence detection head as well. Then, fine-tuning the detection head, in turn, results in a high-accuracy human detection output.

\subsection{SAUL: a curriculum learning strategy}
\label{saulsec}
Still one question remains: how should we \textit{present} the data during training to learn a system which does not predominantly rely on one of the sensing modalities only? Indeed, a naive learning strategy may result in a system that has learned to rely on e.g., the RGB data only, not taking advantage of the other modalities, which is fatal in the case of a hard RGB sensor fault. To address this issue, the authors in \cite{origCrossfusion} introduced a procedure called \textit{BlackIn}, where each modality $j$ of the input data has a probability $p_j$ of being zeroed out during training. In our case, the input tensor is the concatenation of the three RGB channels, the two DVS channels and the radar channel (\ref{tenssse}). Using \textit{BlackIn}, the input training example $n$ can be written as:
\begin{equation}
    T_n^{b} = (\mathcal{B}_{p_1}I_n^{3},\mathcal{B}_{p_2}D_n^{2},\mathcal{B}_{p_3}R_n^{1})
    \label{prune}
\end{equation}
where $\mathcal{B}_{p_j}$ is a Bernoulli-distributed random variable with probability $p_j$. 
Here, we remark that the stochasticity introduced by \textit{BlackIn} is \textit{stationary} during training as the $p_j,\forall j$ are fixed during the training procedure. In contrast, we use \textit{curriculum learning} \cite{curr} that relies on a changing dataset during training 
(a non-stationary process). Formally, let $Q_\lambda$ be a sequence of distributions modelling the target training distribution from which the function of interest should be learned, with a content difficulty parametrized through $\lambda \in [0, 1]$. Then, the sequence $Q_\lambda$ is defined as a \textit{curriculum} if \cite{curr}:
\begin{equation}
    H(Q_\lambda) < H(Q_{\lambda+\epsilon}) \hspace{10pt} \forall \epsilon > 0
    \label{currdef}
\end{equation}
where $H$ denotes the entropy of the distribution. It has been shown that such learning strategy can significantly enhance the model performance on uni-modal tasks such as image recognition, where gradually increasing the difficulty of the images to be recognized (or equivalently, the entropy of the training distribution) helps to obtain a better model \cite{curr}. 

In our case, instead of defining a curriculum over \textit{data examples}, from easy to hard, why not defining it over \textit{sensors}, from low-to high-entropy \textit{sensory} data and combine this principle with \textit{BlackIn}? In order to test this idea, we define the learning procedure of Algorithm 1 by combining \textit{BlackIn} \cite{origCrossfusion} with \textit{curriculum learning} to \textit{shake up} the basins of attraction of the loss landscape in order to ease convergence towards a better local minimum \cite{curr}. We call this new method \textit{\underline{S}h\underline{a}ke-\underline{u}p \underline{L}earning} (SAUL). The algorithm is described below.

\begin{algorithm}
 \caption{SAUL procedure}
  \label{backwards}
 \begin{algorithmic}[1]
 \renewcommand{\algorithmicrequire}{\textbf{Input:}}
 \renewcommand{\algorithmicensure}{\textbf{Output:}}
 \REQUIRE $\mathbf{T}=T_n^{b}, \forall n$ : training data (\ref{prune}), $\vec{p}_a, \mathcal{E}_1, \vec{p}_b, \mathcal{E}_2$ : probability vectors and corresponding epochs
  \FOR {epochs in $1$ to $N_e$}
  \IF {epochs $\in \mathcal{E}_1$}
   \STATE $\vec{p} \xleftarrow{} \vec{p}_a$ \COMMENT{initial probabilities of Eq. (\ref{prune})}
   \ELSIF{epochs $\in \mathcal{E}_2$}
   \STATE $\vec{p} \xleftarrow{} \vec{p}_b$ \COMMENT{abrupt change in probabilities of Eq. (\ref{prune})}
  \ENDIF
  \STATE Train$(\mathbf{T},\vec{p})$ \COMMENT{training with pruned dataset according to Eq. (\ref{prune}) where $p_j$ are the components of the $\vec{p}$ vector.}
  \ENDFOR
 \end{algorithmic} 
 \end{algorithm}

In the case of our sensor suite, we initialise the probabilities in (\ref{prune}) such that the RGB data is the most pruned (high $p_1$), followed by the DVS data ($p_2 < p_1$). The radar data is never pruned ($p_3=0$). 
Then, we decrease the pruning rate of both the RGB and the DVS data abruptly in epochs $\mathcal{E}_2$, half way through the training procedure. By doing so, 
we impose a sudden jump in the entropy of the training data distribution. This complies with the curriculum definition of (\ref{currdef}) as the RGB data distribution has a higher entropy than the DVS distribution (edge image with an important number of zero-valued pixels compared to non-zero pixels), which has a higher entropy compared to radar (sparse set of detections with even more null pixels than the DVS). In the next section, we compare the performance of our model trained via \textit{BlackIn} against the performance obtained via SAUL and will report a significant boost in detection and bounding box regression performance.

\section{Experimental Results}
\label{s3}

\subsection{Precision-Recall Analysis}
\begin{figure}[t!]
\centerline{\includegraphics[scale=0.7]{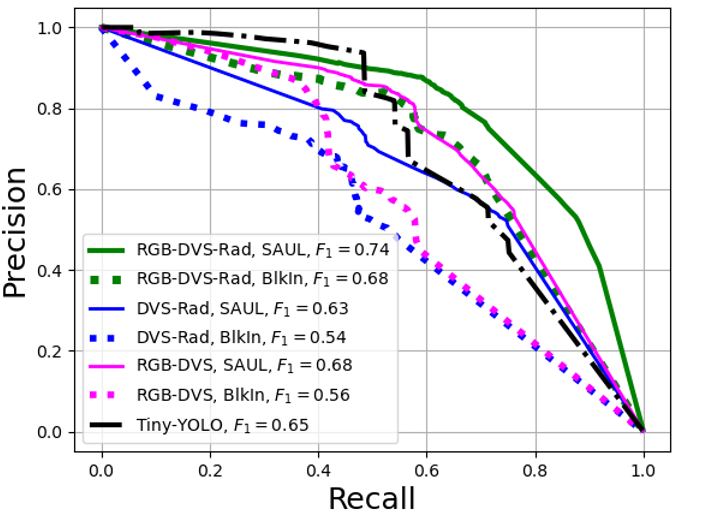}}
\caption{\textit{\textbf{Precision-recall curves} with an IoU threshold of $0.5$. }}
\label{figperf}
\end{figure}

We first train our model for bounding box regression during 300 epochs. We then fine-tune the presence detection head during 40 epochs. For training and testing, we use 5-fold leave-one-out cross-validation as follows. In each cross-validation fold, we keep as test set one fly acquisition with challenging background in Table \ref{acqs} and use the remaining acquisitions as the training set. We use \textit{Adam} with batch size $32$ and learning rate $5\times10^{-5}$ for both training phases. We report the performance of our proposed model by measuring the peak $F_1$ score (i.e., maximum over $i$ of $F_{1,i} = 2 \frac{P_i R_i }{P_i +R_i}$) along each precision-recall curve $(P_i,R_i)$ in Fig. \ref{figperf} (the higher, the better) \cite{averageprec}. During our experiments, we train two models. One model is trained using the \textit{BlackIn} procedure of \cite{origCrossfusion} with $p_1=0.3$ for the RGB and $p_2=0.09$ for the DVS, while keeping $p_3=0$ for the radar in (\ref{prune}). The other model is trained with our proposed SAUL procedure (Algorithm 1) with $\vec{p}_a = [0.7, 0.2, 0]$ (for the first 150 epochs during bounding box learning and the first 20 epochs during detection learning) and $\vec{p}_b = [0.1, 0, 0]$ (for the last 150 epochs and the last 20 epochs of each learning phase), in order to first severely prune the RGB and moderately prune the DVS and then, moderately prune the RGB only. We found those parameter values by testing different combinations and choosing the best ones for both \textit{BlackIn} and SAUL. Doing so, the model is trained to mostly rely on DVS and radar, and to be resilient to an RGB fault since RGB is pruned the most during training. This is in-line with our envisioned scenario where the RGB sensor is considered to be the most fragile modality of the sensor suite. The precision-recall curves are generated for an Intersection over Union (IoU) threshold of $0.5$, by sweeping the threshold of the \textit{presence} output of our CNN (see Fig. \ref{baseCNN}) from $0$ to $1$ \cite{averageprec}. Compared to a pre-trained tiny-YOLO \cite{tinyYOLO}, our cross-fusion system provides a gain of $13\%$ on the peak $F_1$ score. Compared to \textit{BlackIn}, our proposed SAUL procedure provides an average gain of $15\%$ across the different sensor combinations. This clearly shows that using SAUL, our \textit{cross-fusion} model can learn to use the sensing modalities in a significantly better way than through the use of \textit{BlackIn} (achieving PF1 from section \ref{sec:introduction}). Fig. \ref{figperf} also shows the robustness of the system against an RGB fault. Under an RGB sensor fault, a gain of $17\%$ is obtained against \textit{BlackIn} when using our SAUL procedure. In addition, Fig. \ref{figperf} shows the impact of the radar on the peak $F_1$ score. Furthermore, Fig. \ref{figperf} also show that the losses on the precision-recall trade-off due to sensor ablation is more sudden using our proposed SAUL procedure compared to the losses obtained using \textit{BlackIn} (achieving PF4 from section \ref{sec:introduction}). Even though outside our \textit{RGB fault} scenario, it should be noted that the models are very sensitive to a DVS ablation, which is expected since the training procedure mostly relies on DVS and radar at the beginning of the curriculum. 
\subsection{Ablation studies}
\subsubsection{Cross-fusion vs. early-fusion}
\label{crossvsearly}
\begin{figure}[t!]
\centerline{\includegraphics[scale=0.73]{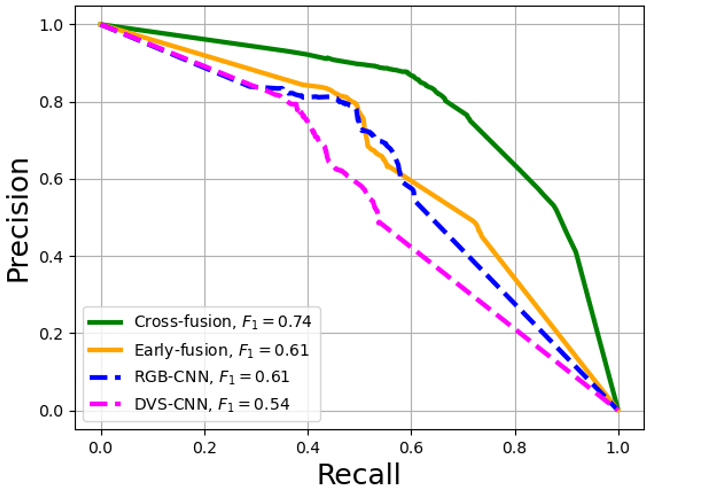}}
\caption{\textit{\textbf{Ablation studies} showing a) cross-fusion vs. early-fusion of RGB, DVS and radar modalities, and b) fusion vs. single modality networks.}}
\label{figperf2}
\end{figure}

In Section \ref{cnnarch}, we motivated our use of a cross-fusion strategy in order to enable the network to learn \textit{at which feature scales} fusion should happen the most. It is therefore important to demonstrate the gain induced by cross-fusion over a traditional \textit{early-fusion} strategy, where all modalities are concatenated once and given as input to the CNN (no cross-fusion highway). Fig. \ref{figperf2} compares the precision-recall curves obtained using our cross-fusion network against early-fusion. The cross-fusion strategy provides a gain of $21\%$ on the peak $F_1$ score. This is expected since the cross-fusion strategy enables the network to learn at which scales the different sensing modalities should be optimally fused.
\subsubsection{Sensor fusion vs. single modality}
In addition, it is important to quantify the robustness gained by fusion of the modalities against single-modality networks. We compare our cross-fusion CNN to both an RGB-CNN and a DVS-CNN. We do not report a radar-trained CNN system since radar detections alone do not convey any features that can be efficiently learned by the CNN (they only indicate the position of a reflector object). Fig. \ref{figperf2} shows the precision-recall curves comparing the single modality networks to our cross-fusion CNN. The cross-fusion strategy provides respective gains of $21\%$ and $37\%$ on the peak $F_1$ score compared to the RGB-only and the DVS-only CNNs.

\subsection{Real-time Performance Analysis}
We demonstrate the real-time feasibility and performance (PF2 and PF3 from section \ref{sec:introduction}) of our system using a Coral edge-TPU as our CNN accelerator (see Fig. \ref{figdr}). We choose the Coral USB accelerator instead of the popular Nvidia Jetson Nano board as the former consumes around $5$ times less power ($\sim 2$W vs. $\sim 10$W) \cite{edgeTPU}, increasing the maximum flight time that the drone can reach (the Raspberry Pi could be later replace by a lower-power system as the bulk of the computation is delegated to the TPU). We deploy our CNN (Fig. \ref{baseCNN}) to the TPU 
using the tools provided in \textit{Tensorflow}. Table \ref{timemarg} reports the mean inference time and standard deviation (evaluated over $10000$ frames) for our CNN running in the edge-TPU. This inference time takes into consideration both the TPU time and the USB communication overhead. A mean inference rate of $77$ FPS is reached, which is more than double the inference rate of a Tiny-YOLO running in similar hardware \cite{tinyYOLO} (this gain could be even more enhanced by exploring modifications on the CNN topology). Naturally, we expect this inference time to slightly grow when using a multi-bounding box output layer \cite{focaloss}, mainly due to the need for non-maximum suppression. Finally, videos showcasing the real-world demonstration of our system are available in the supplementary material provided in the project home page. 
\begin{table}[!t]
\begin{center}
\begin{tabular}{|c|c|}
\hline
\textbf{Mean Inference Time} & \textbf{Standard Deviation}   \\
\hline
13 ms  & 5 ms \\
\hline
\end{tabular}
\caption{
\textit{\textbf{A mean inference rate} of $\mathbf{77}$ \textbf{FPS} is reached (vs. $\sim 30$ FPS for a multi-class Tiny-YOLO \cite{tinyYOLO}).}}
\label{timemarg}
\end{center}
\end{table}
\section{Conclusion}
\label{s4} 
This paper has presented KUL-UAVSAFE, a \textit{first-of-its-kind} dataset fusing RGB, DVS and radar modalities in order to pave the way towards up-most safety and collision avoidance in applications where human agents and drones evolve side-by-side. A baseline CNN architecture trained through our \textit{novel} multi-modal curriculum learning approach (SAUL) has been proposed and compared against \textit{BlackIn}. It has been shown that SAUL provides a significant gain in precision, recall and robustness towards sensor failure, which makes it a well-suited learning strategy for training multi-modal object detector networks.

\end{document}